\documentclass[sn-mathphys,Numbered]{sn-jnl}

\usepackage{graphicx}%
\usepackage{multirow}%
\usepackage{amsmath,amssymb,amsfonts}%
\usepackage{amsthm}%
\usepackage{mathrsfs}%
\usepackage[title]{appendix}%
\usepackage{textcomp}%
\usepackage{manyfoot}%
\usepackage{booktabs}%

\usepackage[plain,lined,shortend,linesnumbered,algo2e,ruled,vlined]
{algorithm2e}
\usepackage{algorithm}%
\usepackage{algpseudocode}
\usepackage{xcolor}%

\usepackage{mdframed}
\usepackage{listings}%
\usepackage{tcolorbox}

\theoremstyle{thmstyleone}%
\theoremstyle{thmstyletwo}%

\theoremstyle{thmstylethree}%

\begin{document}

\title[Article Title]{Geometrically predictable micro fabricated continuum robot}


\author*[1,3]{\fnm{Xiaoyu} \sur{Su}}\email{suxyu@pku.edu.cn}
\equalcont{These authors contributed equally to this work.}

\author[1]{\fnm{Lei} \sur{Wang}}\email{lei\_wang@stu.pku.edu.cn}
\equalcont{These authors contributed equally to this work.}

\author[2]{\fnm{Zhuoran} \sur{Chen}}\email{Ran@mail.nwpu.edu.cn}

\affil[1]{\orgdiv{College of Engineering}, \orgname{Peking University}, \orgaddress{\street{Haidian district}, \city{Beijing}, \postcode{100871}, \country{China}}}

\affil[2]{\orgdiv{School of Astronautics}, \orgname{Northwestern Polytechnical University}, \orgaddress{\street{Youyi West Road}, \city{Xi'an}, \postcode{710072}, \state{Shaanxi}, \country{China}}}

\affil[3]{\orgdiv{School of Mechanical Engineering}, \orgname{Northwestern Polytechnical University}, \orgaddress{\street{Youyi West Road}, \city{Xi'an}, \postcode{710072}, \state{Shaanxi}, \country{China}}}


\abstract{Compared to the micro continuum robots that use traditional manufacturing technology, the micro fabricated continuum robots are different in terms of the application of smart materials, additive manufacturing process, and physical field control. However, the existing geometrical prediction models of the micro continuum robots still follow the model frameworks designed for their larger counterparts, which is inconsistent with the real geometrical transformation principle of micro fabricated continuum robots. In this paper, we present a universal geometrical prediction method for the geometry transformation of the micro fabricated continuum robots based on their material properties and the displacement of the stress points. By discretizing of the micro fabricated continuum structure and applying force constraints between adjacent points to simulate material properties, formulations and simulations are demonstrated to prove the feasibility and effectiveness of the proposed method. Three micro fabricated continuum robots driven through different external field forces are investigated to show two superiorities: the geometrical deformation of a micro fabricated continuum robot under external disturbances can be predicted, and a targeted geometry can be shaped by predicting the sequence and directions of external forces. This pioneer research has contributed to promote understanding and operation of micro fabricated continuum robots and their deformation both from theoretical aspect and real experimental operations.}

\keywords{micro fabricated, continuum robot, geometrical transformation, prediction method}

\maketitle

\section{Introduction}

Owing to the environmental compliance of soft structures and the intelligent controllability of robots, micro continuum robots are well known as a potential technology to move\cite{pancaldi2020flow,heunis2018flexible}, deform\cite{fan2022self,shi2024ultrasound}\cite{yun2012multi,lee2023magnetically,diller2014continuously} and transmit force\cite{fan2024magnetic,kim2022telerobotic} effectively in the intricate and confined pathways of human microvascular networks, which are featured as high tortuosity, multiple bifurcations, and small diameters, thereby enabling a spectrum of minimally invasive medical interventions, including targeted drug delivery, ongoing health diagnostics\cite{lechasseur2011microprobe,hunt2019multimodal}, and telesurgical operations\cite{shi2016shape,russo2023continuum,kim2022telerobotic,peters2016degradable}. Compared with conventional surgical methodologies, the deployment of micro continuum robots is associated with a multitude of clinical benefits, encompassing reduced hemorrhage, diminutive surgical incisions, a decrease in postoperative complications, and expedited patient convalescence\cite{2017Shape}. As an important basis for achieving the above tasks, morphological prediction\cite{pancaldi2020flow,heunis2018flexible,lee2023magnetically,fan2024magnetic,acemoglu2019design} is crucial for the precise movement and deformation of micro continuum robots.

At present, micro continuum robots are usually categorized into two typical paradigms. The first paradigm involves the downscaling of conventional macro scale continuum robots through established manufacturing techniques. Prototypical examples of this category include cable-driven systems with eccentrically arranged cables\cite{liu2022morphology}, hydraulic manipulators with pliable cavity configurations\cite{fan2022self,gopesh2021soft}, concentric tube robots composed of multi-section hollow tubes\cite{peyron2022magnetic}, and snake-shaped robots assembled from interconnected modular elements\cite{wang2024bioinspired}, etc. Limited by the cross-scale mechanical and the complexity of their architectures, the miniaturization of the above robots to sub-millimeter dimensions is impeded, making it difficult to conduct operations within capillary system, which has a wide coverage in human being bodies. The second paradigm is the micro fabricated continuum robot\cite{sahasrabudhe2024multifunctional,garwood2023multifunctional}, such as  the intelligent micro guide wire using hydrogel-doped magnetic materials\cite{2019Ferromagnetic,kim2022telerobotically}, the flexible electronically self-coiling micro catheter\cite{2021Electronically}, and the multimodal magnetron composite soft fiber robot\cite{liu2024magnetic}. As its name implies, micro fabricated continuum robots refers the micro continuum robots that is fabricated through micro manufacturing techniques, such as three dimensional laser direct writing\cite{shang2022customizable,gissibl2016sub}, thin film deposition, LIGA and electrochemical growth. These micro fabricated robots are characterized by a longitudinally uniform structure, which obviates the necessity for multiple cables or modular connections, allowing for deformation at any position along their axis\cite{li2022microfluidic,dreyfus2024dexterous}. In addition, the actuation of these robots is typically mediated by external fields\cite{yang2023magnetically}-either physical or chemical-rather than onboard energy sources\cite{kim2024inherently,zhao2020laryngotors,lee2021non}. The external control force is applied to the robot's stimulus-responsive functional materials at a distance through the physical/chemical field, so as to realize the deformation, motion and micro-manipulation of the micro continuum robots\cite{fu2023magnetically,gu2022artificial,kim2022telerobotic}. In comparison, micro fabricated continuum robots have a smaller diameter, higher degree of freedom, safer, easier to manufacture, and hold considerable promise for clinical application.

Compared to the micro continuum robots that use traditional manufacturing technology, the micro fabricated continuum robots are different in terms of the application of smart materials, additive manufacturing process, and physical field control. However, the existing geometrical prediction models of the micro continuum robots still follow the model frameworks designed for their larger counterparts. For example, joint rotation between multiple rigid bodies is viewed as the basic motion relationship, and the driving force is commonly generated from interactions within the robot's structure. These foundational assumptions of modeling are not consistent with the actual deformation process of micro fabricated continuum robots. So the conventional models cannot accurately reflect the geometrical change process of the micro fabricated continuum robots, whose motion/deformation is driven by physical fields effect on editable materials.

The geometrical modeling of micro fabricated continuum robots diverges from conventional models in three critical respects. Firstly, the micro continuum robots manufactured by conventional technique are often comprised of multi-rigid body modules, and only the relative rotation between modules is studied. but the micro fabricated continuum robot can be bent everywhere, it is more accurate to discretize a micro fabricated continuum structure into mass points for analysis. Secondly, the material properties, which are often overlooked in traditionally manufactured robot's models, are editable for the micro fabricated robots, so the influence of material properties should be factored when calculating micro fabricated continuum robot's geometrical deformation. Thirdly, the driving force of the traditionally manufactured micro continuum robot acts directly on the deformed parts through the structure of the robot itself, thus, the vector information of the external force is obvious and easily-obtained. On the contrary, the driving forces in micro fabricated robots are subject to the spatiotemporal variability of the external physical fields, complicating the real time measurement of forces, and the micro scale manipulation further increases the difficulty of obtaining the force vector. Therefore, the force information of the micro fabricated continuum robot needs to be obtained indirectly by the displacement of the stress point.

\begin{figure}[H]
  \centering
  \includegraphics[width=1\textwidth]{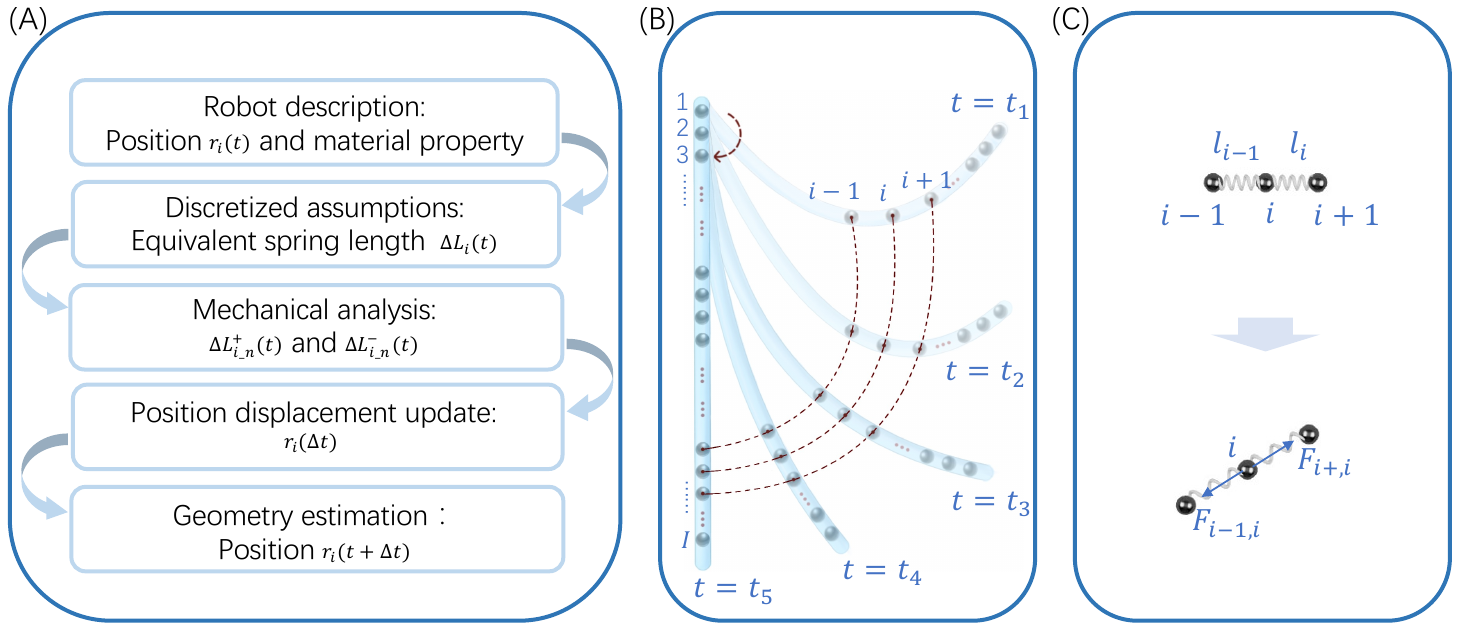} 
  \caption{\textbf{The methodology of the proposed geometrical prediction.} (A): The proposed geometrical prediction method; (B): The sketch of deformation and discretization of a micro fabricated continuum robot; (C): Mechanical analysis between adjacent points.} 
  \label{concept} 
\end{figure}

In this investigation, we propose a universal geometrical prediction method, which can predict robots' geometrical deformation process based on their material properties and the displacement of the stress points(Fig. \ref{concept}(A)). By discretizing the micro fabricated continuum structure into a series of mass points and applying force constraints between adjacent points to simulate material properties, the proposed method has the potential to be applied on a wide variety of micro fabricated continuum robots made of various materials. By this proposed method, the geometrical deformation process of a micro fabricated continuum robot can be predicted under any stresses. Besides, a micro fabricated continuum robot can be shaped into targeted geometry by manipulating the robot according to the proposed method.

The method's efficacy is not only corroborated through Python-based simulations of deformation processes, but also substantiated by two representative experiments, which demonstrate how the proposed method predicts a micro fabricated continuum robot's deformation under unexpected outer disturbance, and how a micro fabricated continuum robot is manipulated into target shape by the strategy planed through our method, respectively. Experimental validation shows a high degree of concordance with the simulated outcomes, indicating the potential of the proposed geometrical prediction method to accurately predict the behavior of micro fabricated continuum robots, marking a significant advance in the field of minimally invasive interventional therapy.

\section{Methodology}\label{sec2}

The proposed methodology is demonstrated based on a general framework of micro fabricated continuum robots. The physical characteristic of a micro fabricated continuum robot is generally defined by the following conditions: the material are uniformly distributed along the long axis, the mechanical properties are consistent everywhere along the long axis, and the length along the long axis is much larger than the dimensionality perpendicular to the long axis. As a robotic structure, a micro fabricated continuum robot has the ability of deforming under control of external fields.

\subsection{Conditions and Assumptions}\label{subsec2}

For practical considerations, we define the situation that a micro fabricated continuum robot is placed on the surface of a microscope stage, where the precise interaction between the robot and the surface is unknown. The robot is controlled by outer physical/chemical fields. The complete shape deformation process could be observed and recorded through a microscope's CCD camera. The position and the displacement of stress points could be accurately measured accordingly.

As is demonstrated in Fig. \ref{concept}(B), the structure of a micro fabricated continuum robot can be simplified into a line segment when it is analyze in the long axis direction. The line segment can be further discretized into a set of mass points, which are constrained to each other by the continuity of the line. The force constraints between adjacent points along the continuum structure can be described by the elastic restraint of a spring(see Fig. \ref{concept}(C)). 

\subsection{Theoretical Formulation.}

\subsubsection{Discretization of a micro fabricated continuum robot.}\label{subsubsec2}

Based on the above conditions and assumptions, a micro fabricated continuum robot in any initial shape can be represented by a set of mass points marked from $1$ to $I+1$ in sequence. The position coordinates corresponding to the point $i(i\in 1,2,3,...,I+1)$ at time $t$ are denoted as $(x_{i}(t),y_{i}(t))$ in a two-dimensional Cartesian coordinate system. The displacement of mass point $i$ at time $t$ is referred as $r_{i}(t)$. 

Therefore, at time $t+\Delta t$, the displacement of a mass point $i$ on the micro fabricated continuum robot could be given as $r_{i}(t+\Delta t)$. Expand by Taylor formula (keep items with only 2 highest order), $r_{i}(t+\Delta t)$ can be expressed by $r_{i}(t)$ and other items related to $r_{i}(t)$:

\begin{equation}\label{1}
r_{i}(t+\Delta t)=r_{i}(t)+r_{i}' (t)\cdot \Delta t+ \frac{{r}''_{i}(t)}{2} \Delta t^{2}   
\end{equation}

Similarly, the displacement of point $i$ at time $t-\Delta t$ could written as:

\begin{equation}\label{2}
r_{i}(t-\Delta t)=r_{i}(t)-r_{i}' (t)\cdot \Delta t+ \frac{{r}''_{i}(t)}{2} \Delta t^{2}   
\end{equation}

Substituting \eqref{2} into \eqref{1}, the displacement $r_{i}(t+\Delta t)$ can be re-written as:

\begin{equation}\label{7}
r_{i}(\Delta t)=r_{i}(t)+{r}''_{i}(t) \Delta t^{2}-r_{i}(t-\Delta t)
\end{equation}

Equation\eqref{7} means that, for any mass point in a micro fabricated continuum robot, a slight displacement after time $t$ can be calculated by the already-known displacement information before time $t$. 

\subsubsection{Kinetic process of a mass point.}

It is assumed that the force constraint between point $i$ and point $i+1$ at time $t$ is equivalent to a spring constraint of length $l_{i}(t)(i\in 1,2,3,...,I)$(Fig. \ref{concept}(C)). At time $0$, $l_{i}(t)$ is noted an equivalent spring original length $l$. Once one stress point is forced to move when an outside disturbance is applied to the micro fabricated continuum robot, the movement of that stress point would cause the change of locations of its adjacent points. After being forced to motivated at time $t=0$, point $i-1$ in the micro fabricated continuum robot would move away from point $i$. At time $t$($t>0$), $l_{i}(t)$ becomes larger than the original length $l$, that is $l_{i-1}(t)>l$. Meanwhile, point $i$ is moving away from point $i+1$, therefore $l_{i}(t)$ is also larger than $l$($l_{i}(t)>l$). So point $i$ is pulled by point $i+1$ and point $i-1$ in the opposite directions at the same time. 

We refer the pull force of point $i-1$ to point $i$ as $F_{i-1,i}(t)$, and the pull force of point $i+1$ to point $i$ as $F_{i+1,i}(t)$. Thus, the resultant force applied on point $i$ can be expressed as follows:

\begin{equation}\label{5}
F_{i}(t)=F_{i-1,i}(t)+F_{i+1,i}(t)
\end{equation}
where
\begin{equation}
 \left \{
  \begin{aligned}
F_{i-1,i}(t)=k\cdot \Delta l_{i-1}(t)\\
F_{i+1,i}(t)=-k\cdot \Delta l_{i}(t)
  \end{aligned}
  \right.
\end{equation}
where $k$ is the elastic coefficient of the spring, which can be calculated through $ k=\frac{E}{2(1+\lambda)}$, $E$ and $\lambda$ is the Young's modulus and Poisson's ratio, respectively. So $k$ is a typical material property in our method.

According to Newton's second law, the accelerated velocity of point $i$ at time $t$ can be expressed explicitly by $F_{i}(t)$ and the mass of point $i$:

\begin{equation}\label{6}
{r}''_{i}(t)=\frac{F_{i}(t)}{A\cdot L_{i}(t) \cdot \rho}
\end{equation}

where $A$ and $\rho$ are the area and density of the cross-section of the micro fabricated continuum robot, respectively. $L_{i}(t)$ means the length near point $i$, calculated by $L_{i}(t)=(l_{i-1}(t)+l_{i}(t))/2$. 

Considering that $r_{i}(t)\approx r_{i}(t+\Delta t)\approx r_{i}(t-\Delta t)$ when $\Delta t \rightarrow0$, $r_{i}(\Delta t)$ can be expressed by substituting \eqref{6} into \eqref{7}:

\begin{equation}\label{14}
  \begin{aligned}
r_{i}(\Delta t)&=\frac{F_{i}(t)}{A\cdot L_{i}(t) \cdot \rho} \cdot \Delta t^{2}\\
&=c\cdot \Delta L_{i}(t)\cdot \Delta t^{2}
  \end{aligned}
\end{equation}

Mark the direction that is consistent with the direction of the displacement of point $i$ as $+$, mark the direction that is opposite to the direction of the displacement of point $i$ as $-$, then the distance between point $i$ with its neighbor points $i-1$ and $i+1$ are obtained by:

\begin{equation}
 \left \{
  \begin{aligned}
  \Delta L^{+}_{i}(t)&=(r_{i-1}(t)-r_{i}(t))\cdot (\frac{\overrightarrow {r_{i-1}(t)r_{i}(t)}}{\left |r_{i-1}(t)r_{i}(t)  \right | } )  \\
    \Delta L^{-}_{i}(t)&=(r_{i}(t) + \Delta L^{+}_{i}(t) -r_{i+1}(t))\cdot (\frac{\overrightarrow {r_{i+1}(t)(r_{i}(t)+\Delta L^{+}_{i}(t))}}{\left |r_{i+1}(t)(r_{i}(t)+\Delta L^{+}_{i}(t))  \right | } )  
    \end{aligned}
    \right.
\end{equation}

\subsubsection{Discretization of the displacement of mass points.}

When $r_{i}(t+\Delta t)-r_{i-1}(t)>l$, the velocity and elastic tension of point $i$ and $i-1$ change with their real time position. Assuming the pulling process of the equivalent spring in each $\Delta t$ is discretized into a set of infinitesimal pulls, then the displacement of point $i$ within $\Delta t$ can be expressed by $N$ parts of displacements. The $n_{th}(n\in 0,1,2...N)$  part of displacement is termed as:

\begin{equation}
    r_{i\_n}(t)=(x_{i\_n}(t),y_{i\_n}(t))
\end{equation}

At time $t$, the $n_{th}$ part of displacement that point $i$ moves towards its neighbor points $i-1$ and $i+1$ are expressed sequentially as:

\begin{equation}\label{8}
 \left \{
  \begin{aligned}
\Delta L^{+}_{i\_n}(t)&= (r_{i-1\_n}(t)-r_{i\_n-1}(t)-l)\cdot \frac{\overrightarrow {r_{i-1\_n}(t) r_{i\_n-1}(t)}}{\left |r_{i-1\_n}(t) r_{i\_n-1}(t)\right | } ) \\
\Delta L^{-}_{i\_n}(t)&= (r_{i\_n-1}(t) + \Delta L^{+}_{i\_n}(t) -r_{i+1\_n-1}(t)-l)\cdot \frac{\overrightarrow {r_{i+1\_n-1}(t) (r_{i\_n-1}(t) + \Delta L^{+}_{i\_n}(t))}}{\left |r_{i+1\_n-1}(t) (r_{i\_n-1}(t) + \Delta L^{+}_{i\_n}(t))\right | } ) 
    \end{aligned}
    \right.
\end{equation}

For points in the end of the micro fabricated continuum robot, such as point $1$ and point $I+1$, the spring pulls only from one side of the points. By substituting \eqref{4} and \eqref{8} into \eqref{14}, we can obtain the following equation:

\begin{equation} \label{9}
    \begin{aligned}
      r_{1\_n}(t+\Delta t)&=r_{1\_n}(t)+r^{+}_{1\_n}(\Delta t)\\
      &=r_{1\_0}(t)+\sum_{j=0}^{n}r_{1\_j}(\Delta t)  \\
      &=r_{1\_0}(t)+\frac{\sum_{j=0}^{n}c\cdot \Delta L_{1\_j}(t)\cdot \Delta t^{2}}{2}     \\
      &=r_{1\_0}(t)
      +\frac{(r_{0\_N^{-}}(t)-r_{1\_0}(t)-l)\cdot c\cdot \Delta t^{2}\cdot (\frac{\overrightarrow {r_{1\_0}(t)r_{0\_N^{-}}(t)}}{\left |r_{1\_0}(t)r_{0\_N^{-}}(t)  \right | } ) }{2}\\
      &+\frac{\sum_{j=1}^{n}[(r_{0\_N^{-}}(t)-r_{1\_j}(t+\Delta t)-l)\cdot c\cdot \Delta t^{2}\cdot (\frac{\overrightarrow {r_{1\_j}(t)r_{0\_N^{-}}(t)}}{\left |r_{1\_j}(t)r_{0\_N^{-}}(t)  \right | } )] }{2}  
    \end{aligned}
\end{equation}

As for any point $i\in [2,I]$, the equivalent spring force pulls from both sides besides $i$. Consequently, the $n_{th}$ part of displacement of point $i$ at time $t$ can be expressed as:

\begin{equation}\label{jia}
    \begin{aligned}
        r_{i\_n}(t+\Delta t)&=r_{i\_n-1}(t)+r^{+}_{i\_n-1}(\Delta t)-r^{-}_{i\_n-1}(\Delta t)
    \end{aligned}
\end{equation}

where

\begin{equation}\label{10}
 \left \{
    \begin{aligned}
        r^{+}_{i\_n-1}(\Delta t)&=\frac{c\cdot \Delta L^{+}_{i\_n-1}(t)\cdot \Delta t^{2}}{2}     \\
        &=\frac{(r_{i-1\_n}(t)-r_{i\_n-1}(t)-l)\cdot c\cdot \Delta t^{2}\cdot (\frac{\overrightarrow {r_{i\_n-1}(t)r_{i-1\_n}(t)}}{\left |r_{i\_n-1}(t)r_{i-1\_n}(t)  \right | } ) } {2}\\  
        r^{-}_{i\_n-1}(\Delta t)&=\frac{c\cdot \Delta L^{-}_{i\_n-1}(t)\cdot \Delta t^{2}}{2}     \\
        &=\frac{(r_{i\_n-1}(t)+r^{+}_{i\_n-1}(\Delta t)-r_{i+1\_n-1}(t)-l)\cdot c\cdot \Delta t^{2}\cdot (\frac{\overrightarrow {r_{i+1\_n-1}(r_{i\_n-1}(t)r^{+}_{i\_n}(\Delta t))}}{\left |r_{i+1\_n-1}(r_{i\_n-1}(t)r^{+}_{i\_n}(\Delta t)) \right | } ) } {2}\\  
        &(n=1,2,3...N)
    \end{aligned}
    \right.
\end{equation}

Through equation \eqref{jia}, the $n_{th}$ part of displacement of point $i$ at time $t$ can be expressed as follows:

\begin{equation}  \label{rRequation}
    \begin{aligned}
        r_{i\_n}(t+\Delta t)&=r_{i\_0}(t)+\sum_{j=0}^{n-1}r^{+}_{i\_j}(\Delta t)-\sum_{j=0}^{n-1}r^{-}_{i\_j}(\Delta t)  \\
        &=r_{i\_0}(t)+\frac{\sum_{j=0}^{n-1}c\cdot \Delta L^{+}_{i\_j}(t)\cdot \Delta t^{2}}{2}-\frac{\sum_{j=0}^{n-1}c\cdot \Delta L^{-}_{i\_j}(t)\cdot \Delta t^{2}}{2}     \\
        &=r_{i\_0}(t)+\frac{\sum_{j=0}^{n-1}[(r_{i-1\_j+1}(t)-r_{i\_j}(t)-l)\cdot c\cdot \Delta t^{2}\cdot (\frac{\overrightarrow {r_{i\_j}(t)r_{i-1\_j+1}(t)}}{\left |r_{i\_j}(t)r_{i-1\_j+1}(t)  \right | } )]  }{2}  \\
        &-\frac{\sum_{j=0}^{n-1}[(R_{i\_j}(t)-r_{i+1\_j}(t)-l)\cdot c\cdot \Delta t^{2}\cdot (\frac{\overrightarrow {r_{i+1\_j}(t)R_{i\_j}(t)}}{\left |r_{i+1\_j}(t)R_{i\_j}(t)  \right | } )]  }{2} 
    \end{aligned}
\end{equation}

where

\begin{equation}
    \begin{aligned}
        R_{i\_j}(t)=r_{i\_0}(t)
        +\frac{\sum_{k=0}^{j-1}[(r_{i-1\_k+1}(t)-r_{i\_k}(t)-l)\cdot c\cdot \Delta t^{2}\cdot (\frac{\overrightarrow {r_{i\_k}(t)r_{i-1\_k+1}(t)}}{\left |r_{i\_k}(t)r_{i-1\_k+1}(t)  \right | } )]  }{2}  
    \end{aligned}    
\end{equation}

When $n=N$, $r_{i\_n}(t+\Delta t)$ is in fact the displacement of point $i$ at time $t$, so the position change of any point between time $t$ and $t+\Delta t$ in a micro fabricated continuum robot can be calculated by equation \eqref{rRequation}. Since the location of every point $i(i\in 1,2,3,...,I+1)$ at time $t$ is known, the new position of every point $i$ at time $t+\Delta t$ can be obtained. The the shape deformation of a micro fabricated continuum robot is then figured out by sequentially updating every points' position. A pseudocode is put here for a general illustration of the whole algorithm.

\begin{tcolorbox}
[colback=blue!10!white,colframe=black,colbacktitle={blue!20!},boxrule=0pt,toprule=0pt,bottomrule=0pt,rightrule=0pt,coltitle={black},fonttitle=\bfseries,title={Algorithm: Geometrical prediction of micro fabricated continuum robots}]
	\SetAlgoLined
	\SetKwInput{Input}{Input}
	\SetKwInput{Output}{Output}
	\Input{$r_{i}(t);k$}
	\Output{$r_{i}(t+ \Delta t)$}
Initialization:$I \leftarrow 30$ , $l \leftarrow 5$, $N \leftarrow 10$, $\Delta t \leftarrow 0.1$\\
 \While{$j<N$}
 {
  \For{$i=1$ \KwTo I}
  {
                \If{$|r_{i\_j}(t)-r_{i+1\_j-1}(t)|>l$}
                {
                    \eIf{$i==1$}
                    {
                    $r_{i\_j}(t+\Delta t)=r_{i\_j-1}(t)+r^{+}_{i\_j-1}(\Delta t)$ 
                    }
                    {
                    $r_{i\_j}(t+\Delta t)=r_{i\_j-1}(t)+r^{+}_{i\_j-1}(\Delta t)-r^{-}_{i\_j-1}(\Delta t)$
                    }
                }
  }
  $j \leftarrow j+1$
 }

\end{tcolorbox}

\subsection{Material and manufacturing}	

For the experimental demonstration of our geometrical prediction methodology, micro fabricated continuum robots are manufactured with standard components and process flow, in order to control its structural mechanics precisely through changing the material properties and dimensional parameters. 

\subsubsection{Preparations of Materials}
Acrylic ester (AAm, 99\%), N-isopropylacrylamide (NIPAAm, 98\%), ethyl lactate (EL, 98\%), polyvinylpyrrolidone (PVP, average molecular weight, ~1,300,000), N,N-dimethylformamide (DMF, 99.5\%), 4,4’-bis(diethylamino)benzophenone (EMK, 97\%), Dipentaerythritol pentaacrylate (DPEPA, 98\%), and triethanolamine (TEA, 99\%) were purchased from Aladdin Chemicals. All chemicals were used directly as received.

\subsubsection{Preparation of hydrogel precursor}
In a typical procedure, NIPAAm and AAm were added to EL solution and stirred vigorously for 30 min. PVP was added to adjust the photoresist viscosity to Adjust the density of the cross-linked network in the material after polymerization. After complete dissolution, the above solution, DPEPA cross-linker, TEA photosensitizer, and the EMK/DMF photoinitiator solution were mixed and stirred overnight to ensure the complete mixing of each component in the solution. Once made, the photoresist was stored in yellow light condition to prevent unnecessary exposure before use. 

\subsubsection{Fabrication of micro continuum robot} \label{fabrication}
In this section, the fabrication procedure of a robot, which is mounted with a nickel magnetic bar, is provided as an example of the standard components and process flow. 

Micro continuum robots were fabricated by two-step sequential printing, using a commercial two-photon direct laser writing system (Nanoscribe GmbH, Germany). 3D models were built using the AutoCAD software and exported as STL format files. The STL files were converted into General Writing Lauguage by the Describe software and loaded into the Nanoscribe software for printing.
Before the first printing, the borosilicate glass square substrate (22 mm by 22 mm, 0.13 to 0.17 mm thick; Thermo Fisher Scientific Inc.) was rinsed using isopropanol, followed by nitrogen purging and oxygen plasma treatment, to serve as the printing substrate. A nickel magnetic bar is positioned at the center of the substrate. Subsequently, a drop of hydrogel precursor was placed on the center of the substrate using a pipette, which was printed with a commercially available 3D laser lithography system (Photonic Professional GT, Nanoscribe GmbH) with a 63×/1.4 oil immersion objective (Zeiss). During the first printing, the laser power (40 mW) and scanning speed ("4000 $\mu$m/s" ) were set to cross-link the hydrogel. The laser power and scanning speed were modulated dynamically to achieve desired stiffness and hardness. After the first printing, the sample was developed in IPA two times for 8 minutes and then dried under nitrogen. After the first fabrication, the nickel magnetic bar is connected to the flexible part.
The IP-L resist was dropped on the existing structure on the glass substrate, which was placed back in the DLW system in the same position as the first printing. Before the second printing, the center of the printing stage was adjusted with the NanoWrite software through the integrated camera in the Nanoscribe to align with the existing structure. During the second printing, the laser power (40 mW) and scanning speed ("8000 $\mu$m/s" ) were set to cross-link the IP-L resist. After the second printing, The sample was immersed separately in SU-8 developer and IPA for 8 minutes and 5 minutes, respectively, for development, and then dried under nitrogen. After the second fabrication, the nickel magnetic bar, the flexible part, and the rigid part are connected together. The micro continuum robots are immersed in DI water for at least a half hour before testing.

\section{Numerical analysis results}

\begin{figure}[H]%
\centering
\includegraphics[width=1\textwidth]{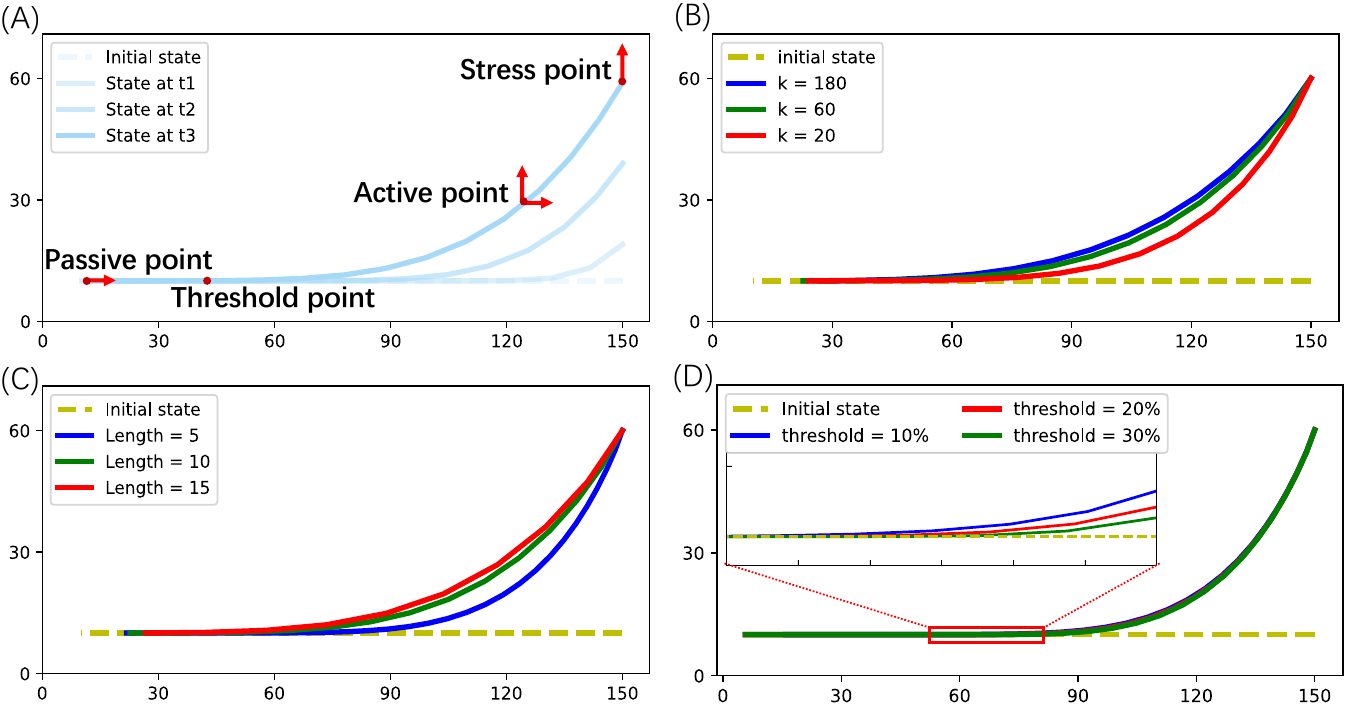}
\caption{Simulated shapes of micro fabricated continuum robots calculated by the proposed method. (A): The geometrical transformation process of a micro fabricated continuum robot predicted by the proposed method; (B): Shape of micro fabricated continuum robots with different elastic coefficient of the spring; (C): The effect of different equivalent spring original length on the geometry of a micro fabricated continuum robot; (D): The effect of different thresholds of the tensile deformation amplitude of the equivalent spring on the geometry of a micro fabricated continuum robot. The unit is $\mu$m in this coordinate system.}
\label{re_png}
\end{figure}

Four simulations are conducted using the libraries NumPy 1.21.5 and Pygame 2.5.2 in Python (Version: Python 3.7.16)(Fig. \ref{re_png}) in this section. Firstly, the geometry deformation process of a micro fabricated continuum robot is simulated to show the computational ability of the proposed method. Then, a simulation is carried out according to our method to evaluate how the material property influences the deformation of a micro fabricated continuum robot. Furthermore, two representative constraints are evaluated in order to analyze the influence factors of the prediction accuracy of the proposed method.

\subsection{Conditions and baseline}

Suppose there is a slender cylindrical micro fabrication robots is placed in the Cartesian coordinate system, whose structure is composed of hydrogel material, with an elastic modulus of $E=60$GPa and a shear modulus of $G=0.2$GPa. In the initial phase, the coordinate of one end of the robot is (10,10), named as the passive point, the other end of the micro fabricated continuum robot is located at (150,10), marked as the stress point. When stress point is subjected to an external force of unpredictable magnitude and direction, the shape of the micro fabricated continuum robot changes. Detailed value of the stress force is hard to measure, but a displacement of the stress point and its nearby points is observed. Considering the above circumstance, which is common in real experiments, the geometrical change of a micro fabricated continuum robot after stress is calculated below by our method.

As a simulation baseline, we discretize that micro fabricated continuum robot as a set of points at a distance of 5$\mu$m along its length direction, the equivalent spring original length is thereupon defined as 5$\mu$m. A stop threshold of the equivalent spring deformation is set in order to prevent the equivalent spring to deforming on and on. When the tensile deformation amplitude of the equivalent spring is greater than the threshold, i.e., $r_{i}(t+\Delta t)-r_{i-1}(t))-l>5\%l$, it is considered that the spring has undergone obvious deformation and produces elastic constraints on both ends. When the tensile deformation amplitude is less than the threshold, i.e., $r_{i}(t+\Delta t)-r_{i-1}(t))-l\leq5\%l$, it is considered that the adjacent points have no relative force constraint and no relative motion. When the tensile deformation amplitude between all points is less than the threshold, it is considered that the form of the micro fabricated continuum robot has completed the deformation and entered a stationary state. 

\subsection{Cases description and analysis}

Given the above conditions, the geometrical deformation of a micro fabricated continuum robot is shown in Figure \ref{re_png}(A). When stress point is forced to move from point (150,10) to (150,60) in the Cartesian coordinate system by an external force, the whole micro fabricated continuum robot is reshaped. Three sequential states during the geometrical deformation process are illustrated in Fig. \ref{re_png}(A), where four points are marked on the state at $t3$. When a point is stressed, the movement of the stress point causes whole structure of the micro fabricated continuum robot to move. The threshold point marks the point achieving the stop threshold of the equivalent spring deformation. To be more specific, the points between the stress point and the threshold point is named as active points, whose nearby equivalent springs deform more then the threshold. Other points are named passive points, which is viewed as unmoved with respect to each other because their nearby equivalent springs deform less then the threshold. Meanwhile, the passive points are not in an absolute static state, instead, they move together towards the stress point. The direction of motion of the active points can be decomposed into two directions towards the stress point, one is consistent with the movement direction of the stress point, the other is perpendicular to the movement direction of the stress point. Comparing the three states at $t1$, $t2$ and $t3$, it is found that both the number of the active points and the displacement of passive points increase along with the movement of the stress point. The displacement distance of active points does not exceed the displacement of the stress point at any states. Meanwhile, the displacement of the point close to the stress point is larger then the displacement of the point away from the stress point. In detail, the magnitude of the displacement produced at each active points is exponential with its distance from the stress point.

The elastic coefficient of the spring $k$ is investigated as one of the most representative parameters of material property. In this simulation, the shape of all three robots are changed by the displacement of stress point from point (150,10) to (150,60). The shapes of three micro fabricated continuum robots are predicted by our method and demonstrated by three lines in Fig. \ref{re_png}(B). The three lines have different curvatures only because of their different value of the elastic coefficient of the spring $k$. The decrease of $k$ means the spring becomes softer, thereupon causes an increase of curvature of the robots' shape.

In order to fully demonstrate the impact of discretization of the micro fabricated continuum robot in our method, the equivalent spring original length and the threshold that produce a tensile deformation are analyzed respectively. 

Firstly, the shape of a micro fabricated continuum robot is calculated by discretizing it into different equivalent spring original length. In Fig. \ref{re_png}(C), the shape of the micro fabricated continuum robot after being forced are separately represented by the  blue, green and red lines, which are denoted as the micro fabricated continuum robots discretized by 5$\mu$m, 10$\mu$m and 15$\mu$m equivalent springs, respectively. It is shown that long equivalent spring original length leads to a worse geometrical continuity of the calculated micro fabricated continuum robots. On the other hand, the stress points of the three lines share the same position, while there is a clear difference between their curvatures, which means the degree of discretization not only effects the continuity of the predicted shape, but also has an influence on the material softness simulation when estimating a geometrical deformation. Comparing Fig. \ref{re_png}(C) and Fig. \ref{re_png}(B), it is found that softer material could be simulated by turning up the equivalent spring original length. 

Secondly, different thresholds of the tensile deformation amplitude of the equivalent spring are simulated in Fig. \ref{re_png}(D). Three lines are separately drawn to represent three micro fabricated continuum robots with $10\%$, $20\%$ and $30\%$ of the threshold. In comparison, the positions where shape change stop in three robots shows a clear sequence in the partial enlargement in Fig. \ref{re_png}(D). To be more specific, the red and yellow robots, representing $20\%$ and $30\%$ threshold sequentially, have shorter reshaped parts with respect to the blue one's. That means when the stress points of the three micro fabricated continuum robots are forced to move to a new location, more discretized points follow up as active points with smaller thresholds. In addition, three robots share the same displacement of stress points, but their smoothness are different. Because smaller threshold allows active points to move more before they turn into inactive state. Therefore, adjacent points can move closer under smaller threshold, so as to form a smoother body shape. Apparently, a smooth geometrical deformation under small threshold is more aligned with real physical phenomenon in the real world, which is consistent with our simulation analysis.
	
\section{Experiment results}

Experiments are designed for the micro fabricated continuum robot manufactured by above process, in order to highlight two main features of the proposed method. 

\subsection{Prediction of geometrical deformation caused by external forces}

A micro fabricated continuum robot is operated under two different situations, in order to prove that: our method is capable to predict how a micro fabricated continuum robot changes its shape when unknown outer disturbances are applied. 

\begin{figure}[H]%
\centering
\includegraphics[width=1\textwidth]{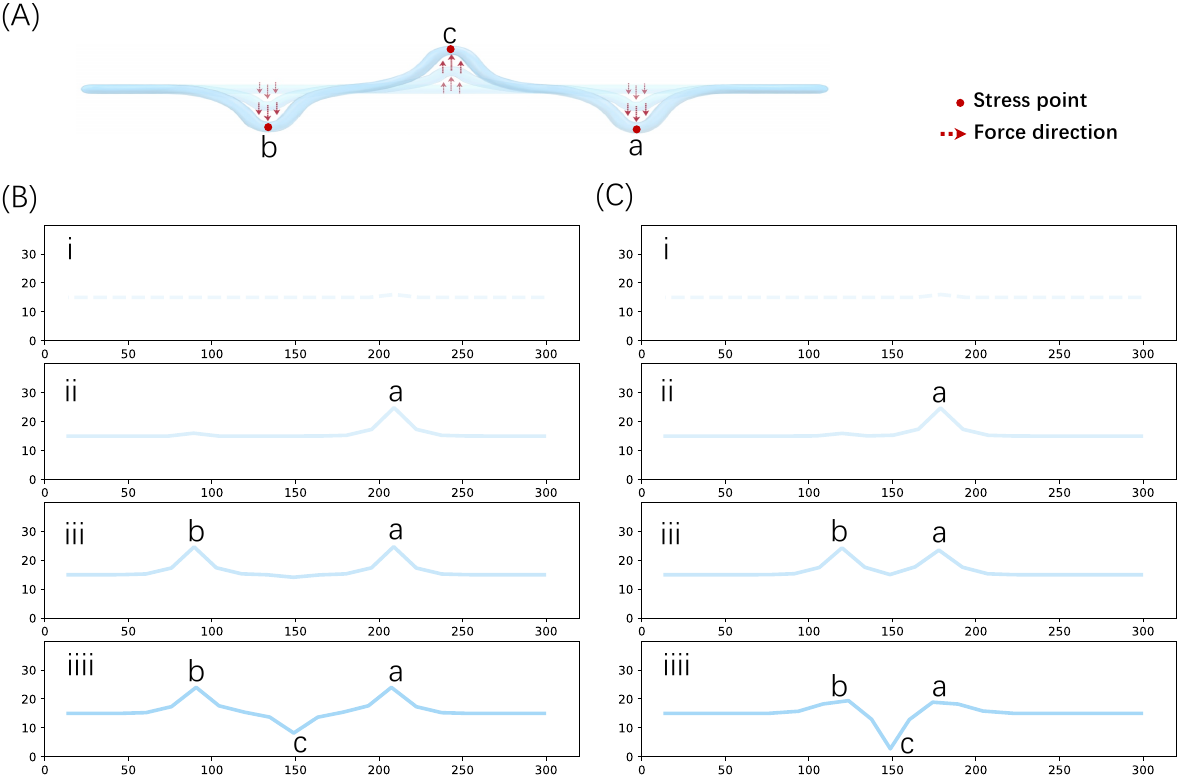}
\caption{Micro fabricated continuum robot with wave like geometrical deformation. (A) An conceptual diagram of geometrical deformation of a micro fabricated continuum robot under external disturbances; (B) A predicting result of geometrical deformation of a micro fabricated continuum robot when stress points distribute at large distance; (C) A predicting result of geometrical deformation of a micro fabricated continuum robot when stress points in close distance. The unit is $\mu$m in this coordinate system.}
\label{wave1}
\end{figure}

As a micro fabricated continuum robot is viewed as a series of points, it is assumed that three forces are sequentially applied vertically on three points of the micro fabricated continuum robot. To be more specific, point a, b and c are forced to move towards the vertical direction with respect to a straight line shaped micro fabricated continuum robot(as is shown in Fig. \ref{wave1}(A)). Among them, point c moves towards the opposite direction with respect to point a and b's direction. Two situations are considered here to verify the proposed method: When the distance between the three points are large, three deformed segments near the three stress points do not effect each other; When the forces are applied on three points whose locations are close enough, the force applied later would cause the whole shape of the micro fabricated continuum robot changes, even though some part of the robot has already been deformed by a previous force. 

The Fig. \ref{wave1}(B) and Fig. \ref{wave1}(C) illustrate the predicted deformation process of a micro fabricated continuum robot under the above two situations, respectively. In both cases, a straight line shaped micro fabricated continuum robot is placed in a two-dimensional Cartesian coordinate system. The positions of those two ends of the robot is (14,15) and (299,15), respectively(see Fig. \ref{wave1}(B)(i) and Fig. \ref{wave1}(C)(i)). 

In the first simulation, put point a, b and c in the position (209, 15), (89,15) and (149,15). When external disturbances come from different directions are applied sequentially on those points, a displacement of 9$\mu$m is observed for each point, meanwhile, independent deformations near the points are predicted by our method(see Fig. \ref{wave1}(B)). In detail, two waves are predicted to form along with the movement of point a and b, towards the upward direction perpendicular to the initial shape of the micro fabricated continuum robot(see Fig. \ref{wave1}(B)(iii)). As the distance between point a and b is 120$\mu$m, while the wave length is only 60$\mu$m around the stress point, the deformation near stress point a and b do not effect each other's wave. It is predicted that disturbance applied on point c causes its displacement in the opposite direction with respect to the previous two waves. Since points a, b and c distribute at a distance of 60$\mu$m with each other, there is an interval of 30$\mu$m between adjacent waves. Therefore, the proposed method predict that three disturbances causes independent geometrical deformation around the three points, and the geometry of the micro fabricated continuum robot is calculated as Fig. \ref{wave1}(B)(iiii). 

In the second simulation, put point a, b and c in the position (179, 15), (119,15) and (149,15). When external disturbances come from different directions are applied sequentially on those points, a displacement of 10$\mu$m is observed for each point, meanwhile, deformations near the points are predicted by our method(see Fig. \ref{wave1}(C)). In detail, two waves are predicted to form along with the movement of point a and b, towards the upward direction perpendicular to the initial shape of the micro fabricated continuum robot(see Fig. \ref{wave1}(C)(iii)). As the distance between point a and b is 60$\mu$m, while the wave length is 60$\mu$m around the stress point, the deformation near stress point a and b do not effect each other's wave. However, when point c is forced to move in the opposite direction with respect to the previous two waves, the deformation of the robot caused by point c's displacement has a severe impact on the geometry formed after the movement of point a and b. The proposed method is capable to predict the final geometry of the micro fabricated continuum robot after the displacement of point c(see Fig. \ref{wave1}(C)(iiii)). According to Fig. \ref{wave1}(C)(iiii), the displacement of point c from (149,15) to (149,2.7) causes the waves around point a and b reshaped. The peaks of the point a and b are dragged along with the movement direction of point c, while the right side of wave b and the left side of wave a are both reshaped and become part of the wave formed by the displacement of point c.

\begin{figure}[H]
\centering
\includegraphics[width=1\textwidth]{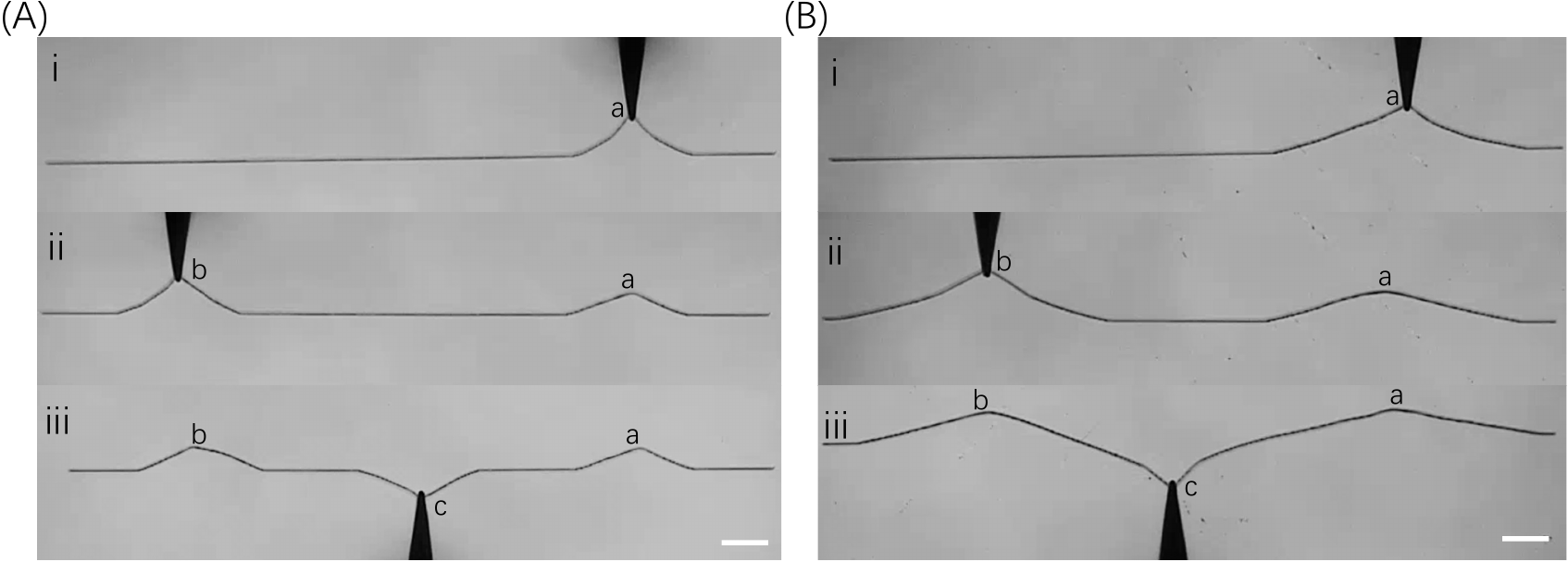}
\caption{Experiment results of micro fabricated continuum robot's geometrical deformation. (A): A experimental result of geometrical deformation of a micro fabricated continuum robot when stress points distribute at large distance; (B): A experimental result of geometrical deformation of a micro fabricated continuum robot when stress points in close distance. The white bar is 200$\mu$m.}
\label{wave2}
\end{figure}

In the above two simulations, we predict a geometrical deformation of a micro fabricated continuum robot from a straight line shape to a wave-like shape, with the only input of the displacement of stress points. Two experiments are carried out under the same conditions, and micro continuum robots are fabricated according to the standard fabrication procedure mentioned in \ref{fabrication}. Two real micro fabricated continuum robots, each of the length is 3$mm$, are pushed by a micro probe to reform their geometry. The deforming results(see Fig. \ref{wave2}) after every pushes are captured by an optical micro scope. 

In the first experiment, point a is moved for a short distance when it is pushed, and a small wave shape is formed around point a, as is shown in Fig. \ref{wave2}(A)(i). Then, point b is pushed by the probe in the same direction and magnitude of the push for point a. It results a new wave near point b as is shown in Fig. \ref{wave2}(A)(ii). After the push of point c, in an opposite direction but the same magnitude as the previous two pushes, the whole structure of the micro fabricated continuum robot is forced to reform in a way the same as the predictions in Fig. \ref{wave1}(B). Compare Fig. \ref{wave1}(B)(iiii) and Fig. \ref{wave2}(A)(iii), it is apparent that the predicted geometry by our method and the experimented robot body is in a high degree of similarity under the circumstance that independent waves are formed by pushing three points.

In the second experiment, geometry transformation of the micro fabricated continuum robot is reshaped by pushing three points with small distances(see Fig. \ref{wave2}(B)). Compared to Fig. \ref{wave2}(A), the distance between point a and b decreases from 2000$\mu$m to 1500$\mu$m, which results in the push on point c reshapes the geometry of the robot after the transformation caused by point a and b. It is shown in Fig. \ref{wave2}(B)(iii) that the experimental geometry transformation is identify with the calculated results in Fig. \ref{wave1}(C)(iiii). It is apparently that the shape of a real micro fabricated continuum robot is almost the same as the shape predicted by the proposed method, which proves that the proposed method has a precise prediction on the geometry of a micro fabricated continuum robot under times of unknown outer disturbances.

\subsection{Shaping a micro fabricated continuum robot into a targeted geometry}

In this section, three different micro fabricated continuum robot are fabricated, aiming to show the other superiority of our method: the sequence and directions of external forces could be deliberately calculated through our method, in order to shape a micro fabricated continuum robot into a targeted geometry. 

\begin{figure}[H]
\centering
\includegraphics[width=1\textwidth]{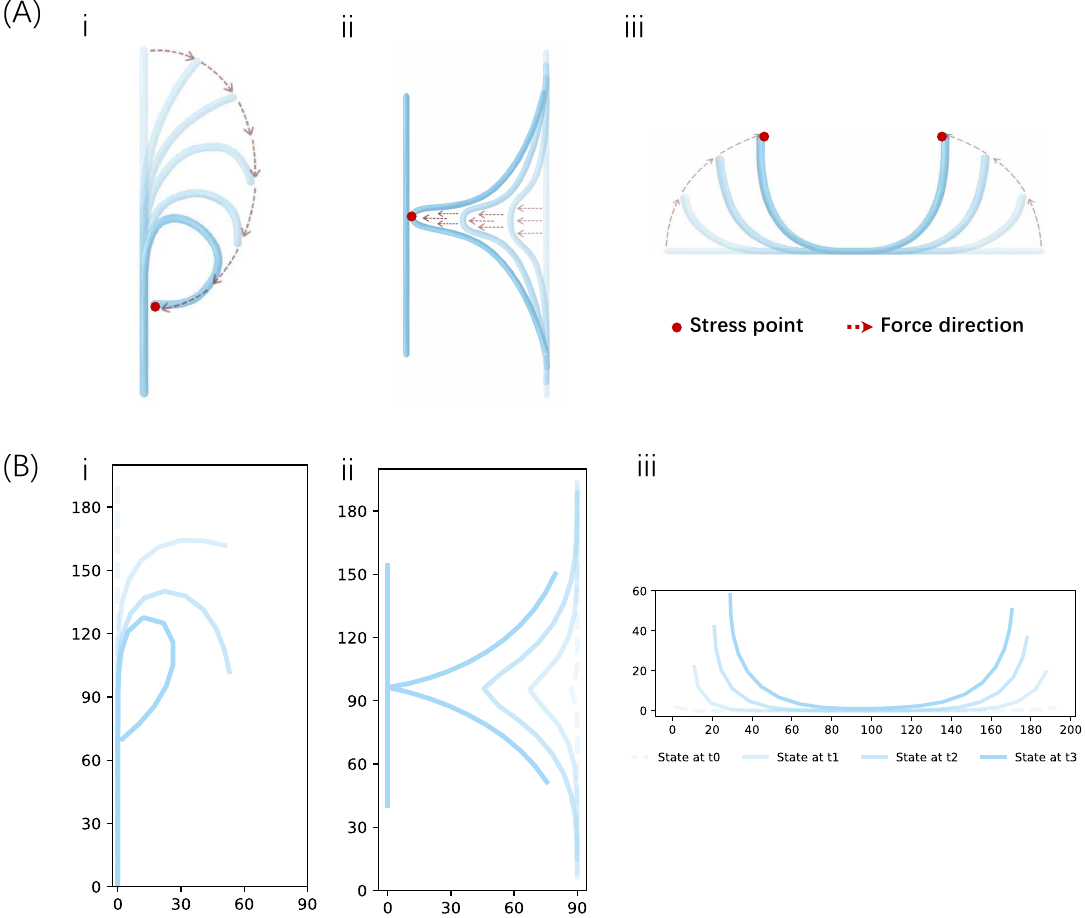}
\caption{Deform micro fabricated continuum robots into designed letters PKU. (A)The concept of geometric deformation process from an initial form of a straight line into the shape of the letter PKU; (B)The predicted geometric deformation into letter PKU by our method}
  \label{Stimuli} 
\end{figure}

In order to verify the effectiveness of the geometrical prediction method proposed in this paper under the action of different external force sources, we use three different external field forces to control the micro fabricated continuum robot to deform into the abbreviation of Peking University, which is PKU(as is shown in Fig. \ref{Stimuli}). In detail, an external force is applied to the micro fabricated continuum robot using a magnetic field, causing it to deform to character P; An acoustic field is used to generate an external force to a micro fabricated continuum robot, deforming it to character K; An external force is applied to a micro fabricated continuum robot through the driving principle of pH stimulus response, deforming it into character U.

\begin{figure}[H]
\centering
\includegraphics[width=0.8\textwidth]{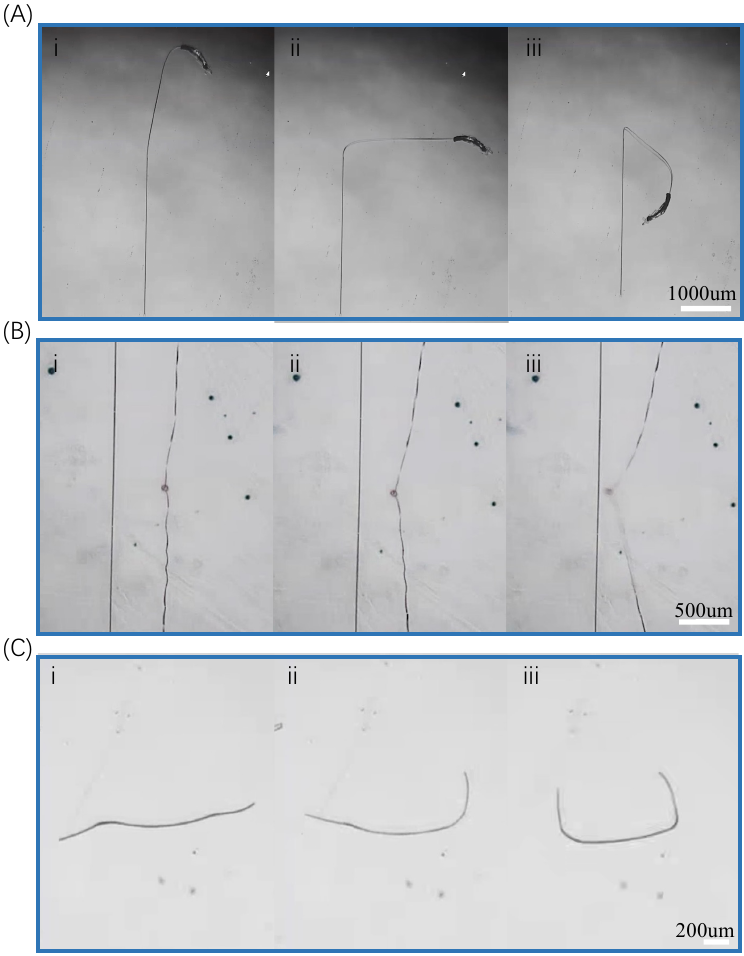}
\caption{The experimental geometric deformation of micro fabricated continuum robots from straight lines to the shape of the letter PKU}
  \label{Exp} 
\end{figure}

In order to use the magnetic field to generate an external force to deform the micro fabricated continuum robot into the shape of the letter P, a continuum robot with an initial form of a straight line is designed. One end of the robot is magnetized as the stress point, so as to make it be able to be dragged by external magnetic field. As is planed, the magnetized stress point of the micro fabricated continuum robot is dragged to move along a way like an arc, until letter P is formed(see the concept in Fig. \ref{Stimuli}(A)(i)). 
To realize that target, our method is used to design a motion path for the magnetized stress point. The initial state of the micro fabricated continuum robot is designed as a straight line between point (0,0) to point (0,190), that is magnetized as the stress point. Then the stress point is designed to move by controllable magnetic field from (0,190) to (51,163), (53,102) and (2,70), as is shown by the four states in Fig. \ref{Stimuli}(B)(i). 
To verify the effectiveness of the designed motion path calculated by our method, a micro fabricated continuum robot is printed according to the fabrication procedure described in \ref{fabrication}, with a bar made of nickle on its end(Fig. \ref{Exp}(A)(i)). Magnetizing the nickel bar with a strong magnetic material and releasing the micro fabricated continuum robot from the coverslip with a probe, so that the micro fabricated continuum robot can be reshaped freely when it is exposed in a magnetic field. Since it is predicted that the shape of a micro fabricated continuum robot would be formed into the letter P through magnetic field force, we place the robot in a three axis Helmholtz coil magnetic field generator which is able to generate a uniform magnetic field of 10$mT$. By controlling the nickel bar of the micro fabricated continuum robot along the path designed by our prediction, the micro fabricated continuum robot is subjected to magnetic traction to form the form of the letter P(as is shown in Fig. \ref{Exp}(A)(iii)). 

Two micro fabricated continuum robots with initial form of straight lines and parallel to each other are fabricated, in order to use the sound field to apply an external force to deform the two robots into the shape of the letter K. A cavity, whose opening is designed in the right side, is fabricated in the middle of the long axis of the right robot, so as to acoustically control the right robot to deform. As is planed, the cavity can be acoustically pushed towards left in liquid environment until the two robots form the letter K(see the concept in Fig. \ref{Stimuli}(A)(ii)). 
To realize that target, our method is used to design a motion path for the magnetized stress point. The initial state of two micro fabricated continuum robots are designed as two straight lines, among which, the right one's ends are point (90,5) and point (90,154). The cavity is designed in point (94,80) as the a stress point. Then the stress point is designed to move by controllable acoustic field from (94,80) to (87,95), (67,95), (46,96) and (0,96), as is shown by the four states in Fig. \ref{Stimuli}(B)(ii). 
To verify the effectiveness of the designed motion path calculated by our method, two micro fabricated continuum robots are printed through micro fabrication technique, with a cavity in the center of the right one(Fig. \ref{Exp}(B)(i)).A ceramic piezoelectric ultrasonic transducer is glued to a glass slide together with a coverslip where the micro fabricated continuum robot is placed, and submerged in pure water mixed with surfactant. A constant-current power supply and amplifier are used to apply a high-frequency alternating voltage to the ultrasonic transducer. When the ultrasonic transducer emits a clear oscillating sound field, due to the oscillation change of sound pressure, the air bubble in the cavity has an oscillating effect on the contact surface between the bubble and the fluid, resulting in the cavity volume on the right micro fabricated continuum robot changes with the sound pressure in pure water. The air bubble in the cavity and the fluid contact surface oscillation produces non-uniform inhalation and jetting force to the fluid in its vicinity, thereby produces an uneven force to the flow field in the surrounding environment. We stimulus the bubble in the cavity, and the middle section of the right continuum robot of micro fabrication is pushed together with the push of the bubble. By controlling the cavity to move along the path designed by our prediction, the micro fabricated continuum robot is subjected to acoustic control to form the form of the letter K(as is shown in Fig. \ref{Exp}(B)(iii)), as is predicted by our method. 

A micro fabricated continuum robot with an initial form of a straight line is fabricated with hydrogel material, in order to use the uneven contraction of the bilayer materials to deform a micro fabricated continuum robot into the shape of the letter U. Two layers of different materials are utilized during fabrication, so as to drive the micro fabricated continuum robot by acid-alkali gradient. As is planed, the robot's body can bend when pH value changes, making the robots to deform into the shape of letter U(see the concept in Fig. \ref{Stimuli}(A)(iii)). 
To realize that target, our method is used to design a motion path for two ends of the robot. The initial state of the micro fabricated continuum robot is designed as a straight line between point (0,0) to point (191,0). In its body, two active areas are made of two layers of materials, whose coefficients of expansion are different. As is planed, the expansion mismatch of the two materials forces the micro fabricated continuum robot to bend asymmetrically into the shape of letter U(see the concept in Fig. \ref{Stimuli}(A)(iii)). 
To verify the effectiveness of the designed motion path calculated by our method, a micro fabricated continuum robot is printed through micro fabrication technique, with two active areas and a large portion of passive area in its body(Fig. \ref{Exp}(C)(i)). To be more specific, two layers of material is fabricated together in the active areas: hydrogel is used to fabricate the pH-driven part and a commercial photoresist (IP-L available from Nanoscribe, Gmbh) is used to fabricate the structural part. Three-dimensional laser direct writer is used to accurately solidify those two materials to construct two bilayer non-uniform shrinkage areas. By this way the micro fabricated continuum robot has the ability to produce shrinkage deformation under acid-base stimulation in its active areas. Due to the mismatch of the shrinkage length of the two layers of materials in the acid-alkali environment, the deformation area of the micro-manufactured continuum robot will bend in the direction of shrinking more, thereby driving the geometry of the whole micro fabricated continuum robot to change. Since it is predicted that the shape of a micro fabricated continuum robot would be formed into the letter P through magnetic field force, we place the robot in a three axis Helmholtz coil magnetic field generator which is able to generate a uniform magnetic field of 10$mT$. By controlling the shrinkage of the active areas of the micro fabricated continuum robot along the path designed by our prediction, the micro fabricated continuum robot is subjected to bend into a letter P shape(as is shown in Fig. \ref{Exp}(A)(iii)). Since it is calculated by the proposed method that the shape of a micro fabricated continuum robot can be deformed into the letter U through acid-alkali gradient as Fig. \ref{Stimuli}(A)(iii), we drop acid solution and alkaline solution to drive the straight line micro fabricated continuum robot deform into the letter U(as is shown in Fig. \ref{Exp}(C)(iii)).

The above experimental demonstrations are all carried out based on the pre-mentioned micro fabricated continuum robot manufactured by standard components and process flow. Nevertheless, various modifications are made to distinguish the three robots, in order to produce stimulus responses to different physical/chemical stimuli without changing the basic structure of the micro fabricated continuum robot. In detail, the first robot is designed with an extra nickel bar, which presents how the proposed method works in a situation that dragging force is applied on a point of the micro fabricated continuum robot. The second robot is designed with an extra air bubble in a cavity, which presents how the proposed method works in a situation that pushing force is applied on a point of the micro fabricated continuum robot. The third robot is designed with an extra material, which presents how the proposed method works in a situation that inner force is applied on a point of the micro fabricated continuum robot. 

In one word, the displacement caused by the above three forces, in spite of the force vectors themselves, are considered as input values when estimating the geometry deformation process in our method. The experimental results (Fig. \ref{Exp}) show that no matter what kind of physical/chemical stimulus is subjected to, and no matter how large the degree of stimulation, as long as the force point is displaced in a specific direction under the stimulus response condition, the overall geometry of the micro fabricated continuum robot also changes due to the continuity of the micro fabricated continuum robot. In micro scale experiments, it is difficult to measure the actual instantaneous forces and moments exerted by the physical field on the robot in a split second due to the very short physical process of the stimulus response, but the camera is able to record and measure the displacement of the stress point. Comparing the simulation (Fig. \ref{Stimuli}(B)) and the experiment (Fig. \ref{Exp}), it can be seen that when a given external force is applied to the upper point of the micro fabricated continuum robot, as long as the displacement input generated by the external force to the stress point is the same, the algorithm proposed in this paper is used to predict the geometrical change process of the micro fabricated continuum robot after stress, which is basically consistent with the actual observation of the micro fabricated continuum robot after the geometrical change process after stress. In micro scale experiments, it is often difficult to accurately measure the external forces of the physical/chemical field on the microstructure test drive, for example, in the experiment in Fig. \ref{Exp}, the force caused by the magnetic moment, acoustic flow and unbalanced contraction of the bilayer structure is not measured, but it is easy to observe the displacement amplitude caused by the external force on the stress point of the microstructure under the microscope. Therefore, the method proposed in this paper can effectively help experimenters to predictably regulate the overall geometry of micro fabricated continuum robots.

\section{Conclusion}

In view of the characteristics of the micro fabricated continuum robot that can be bent everywhere and the material properties can be edited, a basic modeling framework is proposed to simulate the structure of the micro continuum based on discrete particles and simulate the material properties of the micro continuum using the elastic force constraints between the particles. In order to solve the problem that the instantaneous force of the robot is unpredictable in the physical field of micro-scale three-dimensional spatio-temporal variability, a micro robot dynamic model was established by using the easy-to-measure physical quantity of the displacement of the force point combined with the discretization algorithm. Therefore, a geometrical prediction algorithm for continuum robots is established, which is specifically for micro fabrication. According to the algorithm proposed in this paper, the overall geometrical deformation process of the continuum robot can be predicted according to the material characteristic parameters of the micro-manufactured continuous robot by measuring the displacement of the stress point, which is helpful to reduce the number of micro-manipulation physics experiments in the research of micro-manufactured continuum robot and improve the experimental efficiency. Furthermore, by deliberately setting the physical field action mode, the deformation process of the micro fabricated continuum robot can be controlled and turned into a special target form.

In this paper, the effectiveness of the method is verified by cross-comparison of theoretical models and simulation analysis, and micro scale mechanical experiments are adopted as an experimental support. However, this paper focuses on the establishment of a two-dimensional model from the material properties and microstructure parameters of the micro fabricated continuum robot to the geometry of the micro fabricated continuum robot, and conducts experiments and observations in the two-dimensional environment in order to facilitate the characterization in the experiment. Therefore, it will be possible to extend the model in this paper to three-dimensional space in the future, and further verify the model in the geometrical manipulation experiments of three-dimensional micro fabricated continuum robots.

\section{Acknowledgements}
We acknowledge financial support from the Key Projects in Shaanxi Province (No.2024CY2-GJHX-08), the Key Projects in Shaanxi Province (No.2024CY2-GJHX-10), China Postdoctoral Science Foundation (No.2023M740091), and the Foundation from Beijing Innovation Center for Engineering Science and Advanced Technology in Peking University.

\bibliography{sn-bibliography}

\end{document}